\documentclass[conference]{IEEEtran}
\IEEEoverridecommandlockouts
\usepackage{amsmath,amssymb,amsfonts}
\usepackage{textcomp}
\usepackage{times}
\usepackage{helvet}
\usepackage{courier}
\usepackage[hyphens]{url}
\usepackage{graphicx}
\usepackage{cite}
\usepackage{algorithm}
\usepackage{algorithmic}
\usepackage{booktabs} 
\usepackage{multirow} 
\usepackage[caption=false,font=normalsize,labelfont=sf,textfont=sf]{subfig} 
\usepackage{url} 
\usepackage{amssymb} 
\usepackage{pifont} 
\usepackage{newfloat}
\usepackage{listings}
\usepackage{xspace}
\usepackage{hyperref}
\usepackage{microtype}
\usepackage{pdfpages}
\usepackage{bm}
\usepackage{colortbl}
\usepackage{amsmath}
\usepackage{times}
\usepackage{helvet}
\usepackage{courier}
\usepackage{xcolor}
\def\BibTeX{{\rm B\kern-.05em{\sc i\kern-.025em b}\kern-.08em
    T\kern-.1667em\lower.7ex\hbox{E}\kern-.125emX}}
\def\eg{\textit{e.g.}\xspace}
\def\ie{\textit{i.e.}}

\def\name{\emph{DTP}\xspace}
\def\MA{FSD\xspace}
\def\MB{SDR\xspace}
\def\MC{CSR\xspace}
\def\ML{LLISR\xspace}
\def\BibTeX{{\rm B\kern-.05em{\sc i\kern-.025em b}\kern-.08em
    T\kern-.1667em\lower.7ex\hbox{E}\kern-.125emX}}
\begin{document}

\title{Dual-Path Learning based on Frequency Structural Decoupling and Regional-Aware Fusion for Low-Light Image Super-Resolution}
\author{
\IEEEauthorblockN{
Ji-Xuan He\textsuperscript{1,\textdagger},
Jia-Cheng Zhao\textsuperscript{1,\textdagger},
Feng-Qi Cui\textsuperscript{2,*},
Jinyang Huang\textsuperscript{1,*},
Yang Liu\textsuperscript{3},
Sirui Zhao\textsuperscript{2},
Meng Li\textsuperscript{1},
Zhi Liu\textsuperscript{4}
}
\IEEEauthorblockA{
\resizebox{0.96\textwidth}{!}{%
\begin{tabular}{@{}c@{\hspace{1.5em}}c@{}}
\textsuperscript{1}Hefei University of Technology, Hefei, China &
\textsuperscript{2}University of Science and Technology of China, Hefei, China \\
\textsuperscript{3}Zhejiang University, Hangzhou, China &
\textsuperscript{4}The University of Electro-Communications, Tokyo, Japan
\end{tabular}%
}
}
\thanks{\textsuperscript{\textdagger}Equal contribution. \textsuperscript{*}Corresponding authors.}
}
\IEEEaftertitletext{\vspace{-2.2\baselineskip}}
\maketitle
\begin{abstract}

Low-light image super-resolution (\ML) is essential for restoring fine visual details and perceptual quality under insufficient illumination conditions with ubiquitous low-resolution devices.  Although pioneer methods achieve high performance on single tasks, they solve both tasks in a serial manner, which inevitably leads to artifact amplification, texture suppression, and structural degradation. 
To address this, we propose \textbf{Decoupling then Perceive} (\name), a novel frequency-aware framework that explicitly separates luminance and texture into semantically independent components, enabling specialized modeling and coherent reconstruction.
Specifically, to adaptively separate the input into low-frequency luminance and high-frequency texture subspaces, we propose a \textbf{Frequency-aware Structural Decoupling} (\MA) mechanism, which lays a solid foundation for targeted representation learning and reconstruction.
Based on the decoupled representation, a \textbf{Semantics-specific Dual-path Representation} (\MB) learning strategy that performs targeted enhancement and reconstruction for each frequency component is further designed, facilitating robust luminance adjustment and fine-grained texture recovery.
To promote structural consistency and perceptual alignment in the reconstructed output, building upon this dual-path modeling, we further introduce a \textbf{Cross-frequency Semantic Recomposition} (\MC) module that selectively integrates the decoupled representations.
Extensive experiments on the most widely used \ML~benchmarks demonstrate the superiority of our \name framework, improving $+$1.6\% PSNR, $+$9.6\% SSIM, and $-$48\% LPIPS compared to the most state-of-the-art (SOTA) algorithm. Codes are released at \url{https://github.com/JXVision/DTP}. 


\end{abstract}


\vspace{-0.5em}
\section{Introduction}

\begin{figure}[t]
\vspace{-0.5em}
    \centering
    \includegraphics[width=1.05\linewidth]{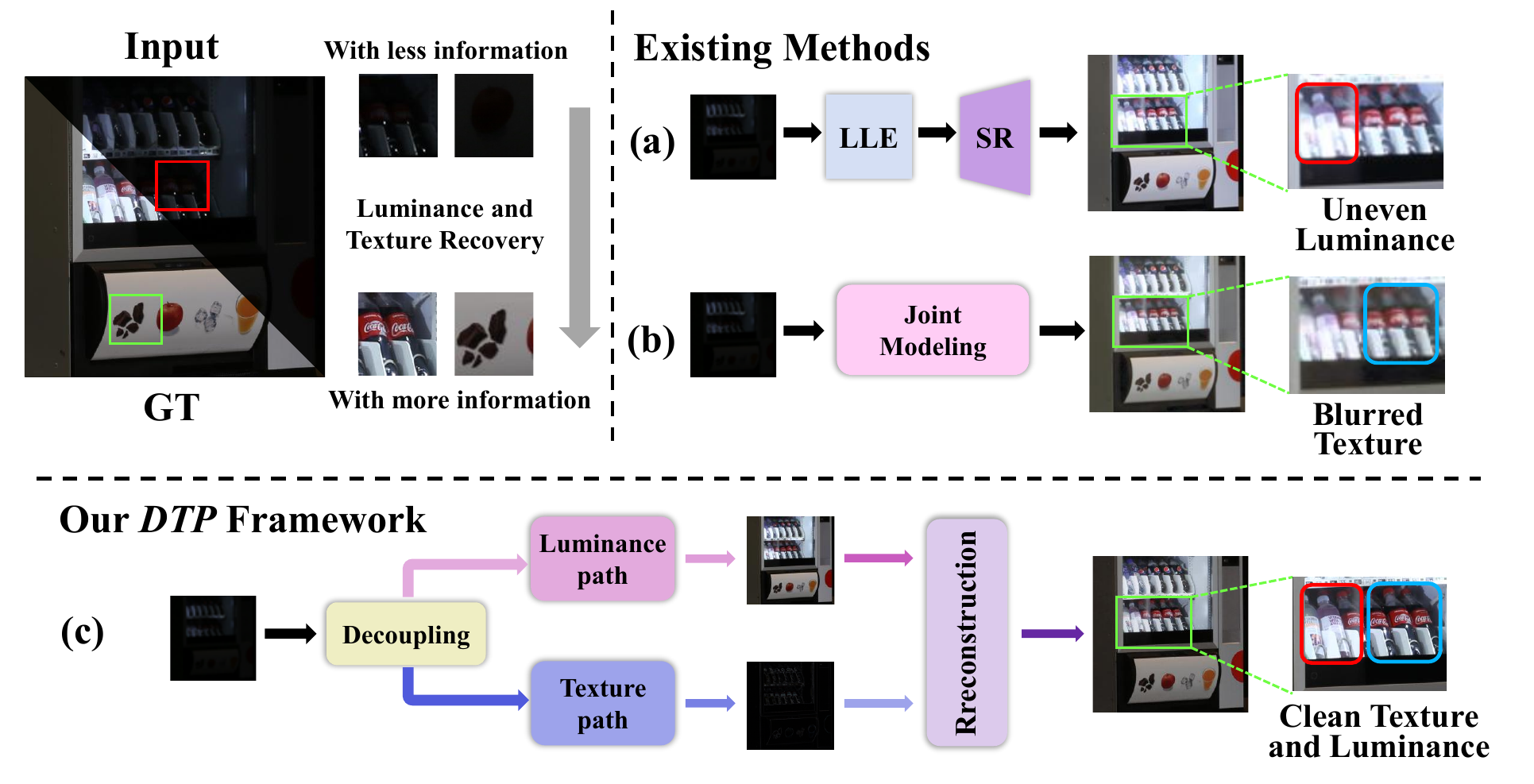} 
    \caption{Comparison of existing methods and our proposed approach. (a) and (b) represent sequential and joint spatial-domain \ML~methods, respectively, which suffer from uneven luminance and blurred textures due to early-stage semantic entanglement. (c) illustrates our frequency-aware disentanglement strategy, which explicitly separates and models luminance and texture components, enabling faithful recovery with both crisp luminance and fine textures.
    }
    \label{fig:motivation}
\vspace{-0.9em}
\end{figure}

Low-light image super-resolution (\ML) is a critical task in computer vision, with applications in night surveillance, autonomous driving, and low-light photography~\cite{aakerberg2021rellisur}. Unlike traditional super-resolution, \ML~must simultaneously address severe luminance degradation and detail loss, making the interaction between luminance and texture a central challenge~\cite{ ye2024learning}.
As shown in Fig. \ref{fig:motivation}, early approaches often adopt a two-stage paradigm that sequentially performs low-light enhancement and resolution reconstruction~\cite{li2021learning}. Although achieving high accuracy, such decoupled designs inevitably suffer from error amplification and structural artifacts due to the error accumulation of the serial structure. To address this, recent methods move toward joint frameworks that integrate enhancement and reconstruction in a unified pipeline~\cite{ yue2024unveiling}, or leverage frequency priors to suppress artifacts~\cite{liu2022noise}. However, these methods still struggle with semantic confusion between luminance and texture, especially under extreme lighting conditions~\cite{li2024perceptual}, which highlights the need for deeper modeling mechanisms and more robust optimization strategies.

While \ML~has achieved remarkable progress, existing methods still struggle in structurally complex regions under severe illumination degradation. A key limitation is their reliance on joint spatial-domain modeling and static fusion, which entangles luminance and texture features and leads to ambiguous semantics, artifact confusion, and degraded perceptual quality.
At the heart of these limitations lie three fundamental challenges:
\textbf{1) Semantic entanglement leads to structural ambiguity.}
Luminance and texture are typically modeled together in a unified feature space, without explicit structural disentanglement. This early-stage coupling blurs semantic boundaries and makes it difficult for the model to distinguish luminance artifacts from true textures. As a result, structural distortion and perceptual inconsistency frequently occur in low-light scenarios.
\textbf{2) Representation lacks semantic selectivity and subspace sensitivity.}
Despite the inherently different perceptual roles and frequency characteristics of luminance and texture, most methods adopt flat or undifferentiated representations, \eg, shared channels, uniform scales, or shallow stacks. This lack of subspace-aware modeling weakens the system’s ability to adaptively respond to different semantic targets, leading to feature interference, insufficient specialization, and unstable representation quality.
\textbf{3) Static fusion fails to adapt to spatial heterogeneity.}
Mainstream fusion strategies often employ fixed weights or naive concatenation, without accounting for regional variations in luminance and texture distribution. Such rigid mechanisms are unable to modulate the integration of heterogeneous features based on content relevance, resulting in semantic misalignment, redundancy, and local structural degradation.

To address the aforementioned challenges, we propose Decoupling to Perceive (\name), a frequency-aware low-light super-resolution framework that explicitly decomposes, models, and fuses heterogeneous visual representations in a structurally guided manner. Specifically, to mitigate early-stage semantic entanglement, a Frequency-aware Structural Disentanglement (\MA) module is first introduced. By adaptively transforming input features, \MA separates perceptually distinct luminance and texture components into low- and high-frequency subspaces, reducing cross-domain interference. Based on the disentangled representation, we design a Semantics-specific Dual-path Representation (\MB) strategy that performs targeted modeling for each subspace. The low-frequency luminance branch enhances brightness perception through a bio-inspired mechanism, while the high-frequency texture branch reconstructs fine structural details using residual learning. This design improves robustness and task-specific expressiveness under low-light conditions.
To further enhance integration, we propose a Cross-frequency Semantic Recomposition (\MC) module that adaptively fuses the decoupled features via spatial-channel attention, enabling structure-aware fusion and perceptual consistency. In summary, these modules synergistically form a unified pipeline that systematically addresses the core limitations of \ML, which further achieves superior fidelity, detail recovery, and perceptual quality across diverse low-light scenarios.

Our main contributions are summarized as follows:
\begin{itemize}
\item{To the best of our knowledge, this paper is the first attempt to introduce the principle of frequency-aware structural disentanglement into \ML. By explicitly decoupling luminance and texture in the frequency domain and aligning modeling pathways with their semantic roles, we construct a unified \name framework that systematically mitigates early-stage semantic entanglement and enhances perceptual consistency in complex low-light conditions.}
\item{A novel frequency-aware disentanglement module \MA is first proposed by adopting an adaptive transformation scheme to separate luminance and texture into distinct frequency subspaces. This design enables early semantic decomposition and produces compact representations for targeted modeling and interference suppression.}
\item{We further introduce a dual-path modeling and adaptive fusion strategy, consisting of \MB and \MC modules. \MB enhances low-frequency and high-frequency components through biologically inspired adjustment and residual refinement, while \MC employs spatial-channel attention to selectively integrate heterogeneous features, improving regional generalization and structural fidelity.}
\item{Extensive experiments on standard \ML~benchmarks validate the effectiveness of \name. Our method consistently outperforms all state-of-the-art approaches across ×2 and ×4 scales, which achieves superior perceptual quality, detail restoration, and generalization robustness in diverse low-light scenarios.}
\end{itemize}

\begin{figure*}[t]
  \centerline{\includegraphics[width=0.9\textwidth]{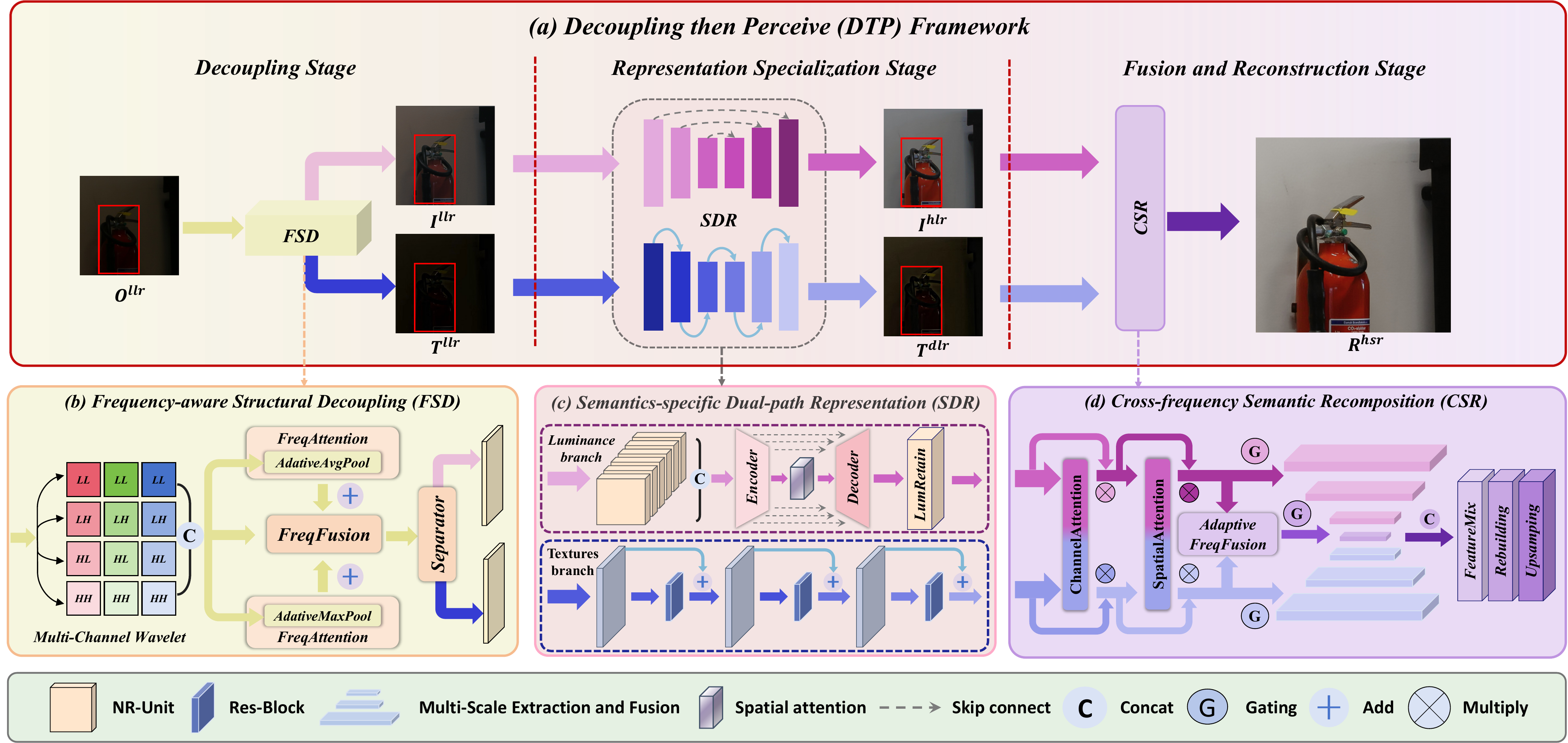}}
  \caption{An overview of the proposed \name (a) The overall processing flow of \name. (b) The pipeline of Decoupling Stage (\MA) in \name. (c) The flow of Representation Specialization Stage (\MB). (d) The sketch of Fusion and Reconstruction Stage (\MC).}

  \label{fig:dtp_pipeline}
\vspace{-10pt}
\end{figure*}

\section{Methodology}
To address semantic ambiguity and structural degradation in \ML, by introducing a structured pipeline of decomposition, specialization, and fusion, we propose a frequency-aware low-light super-resolution framework termed Decoupling then Perceive (\name), whose overall workflow is illustrated in Fig.~\ref{fig:dtp_pipeline}. Specifically, \name first employs a Frequency-aware Structural Decoupling (\MA) module to adaptively separate low-light inputs into low-frequency luminance and high-frequency texture components, mitigating early-stage entanglement. On top of the decoupled representation, a Semantics-specific Dual-path Representation (\MB) strategy enhances brightness perception and reconstructs texture details via two dedicated branches. Finally, a Cross-frequency Semantic Recomposition (\MC) module selectively fuses the processed features through joint spatial and channel attention, promoting semantic alignment and perceptual consistency. 

\subsection{Frequency-aware Structural Decomposition}
To alleviate semantic ambiguity caused by joint spatial modeling\cite{cui2025learning}, we introduce the FSD module to establish a structurally stable representation.

Given a low-light input $O^{llr} \in \mathbb{R}^{H \times W \times 3}$, we first apply a learnable wavelet transform to generate frequency subbands $\{x^{LL}, x^{LH}, x^{HL}, x^{HH}\} = \mathcal{W}_\theta(O^{llr})$, where $x^{LL}$ encodes global luminance and others capture high-frequency textures. To enhance representation consistency, we utilize a learnable vector $\alpha$ (subject to $||\alpha||_1=1$) to reweigh the stacked subbands $f$ via $\hat{f} = \alpha^\top \cdot f$. Furthermore, we constrain the luminance component $f^{LL}$ using a KL divergence loss $\mathcal{L}_{KL}=D_{KL}(P_{LL}||\mathcal{N}(\mu_{0},\sigma_{0}^{2}))$ to ensure compactness. Finally, the fused feature $\hat{f}$ is explicitly separated into independent branches $\{I^{llr}, T^{llr}\}$, which contain structure-preserving low-frequency luminance and fine-grained high-frequency textures, respectively, laying a foundation for targeted processing.

\subsection{Semantics-specific Dual-path Representation }
To fully leverage the separated frequency-aware features in the \MA module, we propose a semantically specific dual-path representation (\MB) module. This module introduces two structurally independent branches that model luminance and texture, respectively, allowing each branch to focus on its respective modality. Specifically, the luminance branch focuses on enforcing global illumination consistency, while the texture branch strives to recover fine-grained structure and suppress noise artifacts. SDR effectively avoids interference between luminance and texture representations. This design enables frequency-specific enhancement while suppressing cross-domain interference, ensuring robust and interpretable representation learning in low-light conditions.

\subsubsection{Luminance Enhancement via Bio-inspired Activation}
In the luminance pathway, we propose a biologically-inspired enhancement mechanism that mimics the adaptive response of human photoreceptors towards varying illumination. Given the luminance component $\mathcal{I}^{llr}$, a nonlinear transformation modeled on the Naka–Rushton equation is applied to enhance the quality of dark regions while avoiding overexposure. The enhanced luminance is computed as:
\begin{equation}
\mathcal{I}^{hlr} = \frac{(\mathcal{I}^{llr})^\gamma}{(\mathcal{I}^{llr})^\gamma + \sigma^\gamma + \beta} \cdot \frac{\mu_{\text{in}}}{\mu_{\text{out}} + \epsilon},
\end{equation}
where $\gamma$, $\sigma$, and $\beta$ control the contrast sensitivity, perceptual threshold, and saturation suppression, respectively. $\mu_{\text{in}}$ and $\mu_{\text{out}}$ denote the mean intensities before and after activation, respectively, and $\epsilon$ is a small constant for numerical stability.

This transformation adaptively scales brightness while maintaining structural contrast, guided by local photometric statistics. Crucially, due to the prior frequency-domain decoupling in \MA, this operation exclusively enhances luminance instead of affecting texture details, thereby preserving semantic integrity and avoiding undesired signal interference.

\subsubsection{Texture Restoration via Residual-based Hierarchical Denoising}
To deal with the high-frequency texture component $\mathcal{T}^{llr}$, which contains rich structural cues but is vulnerable to noise under extreme low-light degradation, we thus introduce a hierarchical denoising framework based on residual learning. The denoising proceeds iteratively, which can be expressed as:
\begin{equation}
\mathcal{T}_i = \mathcal{T}_{i-1} + \mathcal{R}_i(\mathcal{T}_{i-1}), \quad i = 1, \dots, L,
\end{equation}
where we set $\mathcal{T}_0 = \mathcal{T}^{llr}$, and each $\mathcal{R}_i(\cdot)$ denotes a residual unit that selectively removes structured noise while preserving fine-grained details.

This progressive refinement scheme enables stable texture enhancement, balancing high-frequency detail preservation and denoising robustness. After $L$ stages, the final representation $\mathcal{T}^{dlr} = \mathcal{T}_L$ serves as a clean and expressive high-frequency feature fed for subsequent semantic fusion.

\begin{figure*}[t]
  \centerline{\includegraphics[width=0.8\textwidth]{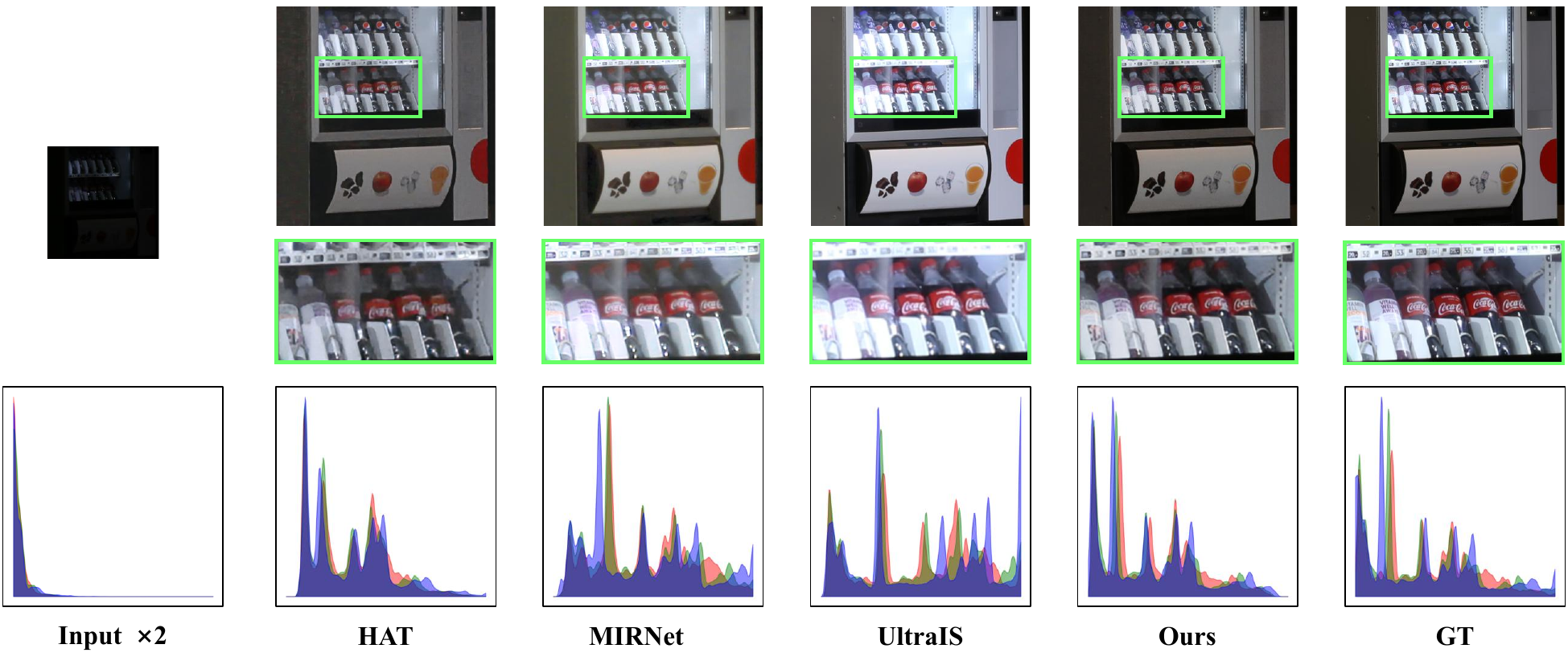}}
  \caption{Qualitative comparisons of $\times$2 tasks on RELLISUR. The top row shows the full restored images, the middle displays zoomed-in patches for local structure inspection, and the bottom presents RGB histogram distributions.}
  \label{fig:RGB}
\vspace{-10pt}
\end{figure*}

\begin{figure}[t]
    \centerline{\includegraphics[width=0.5\textwidth]{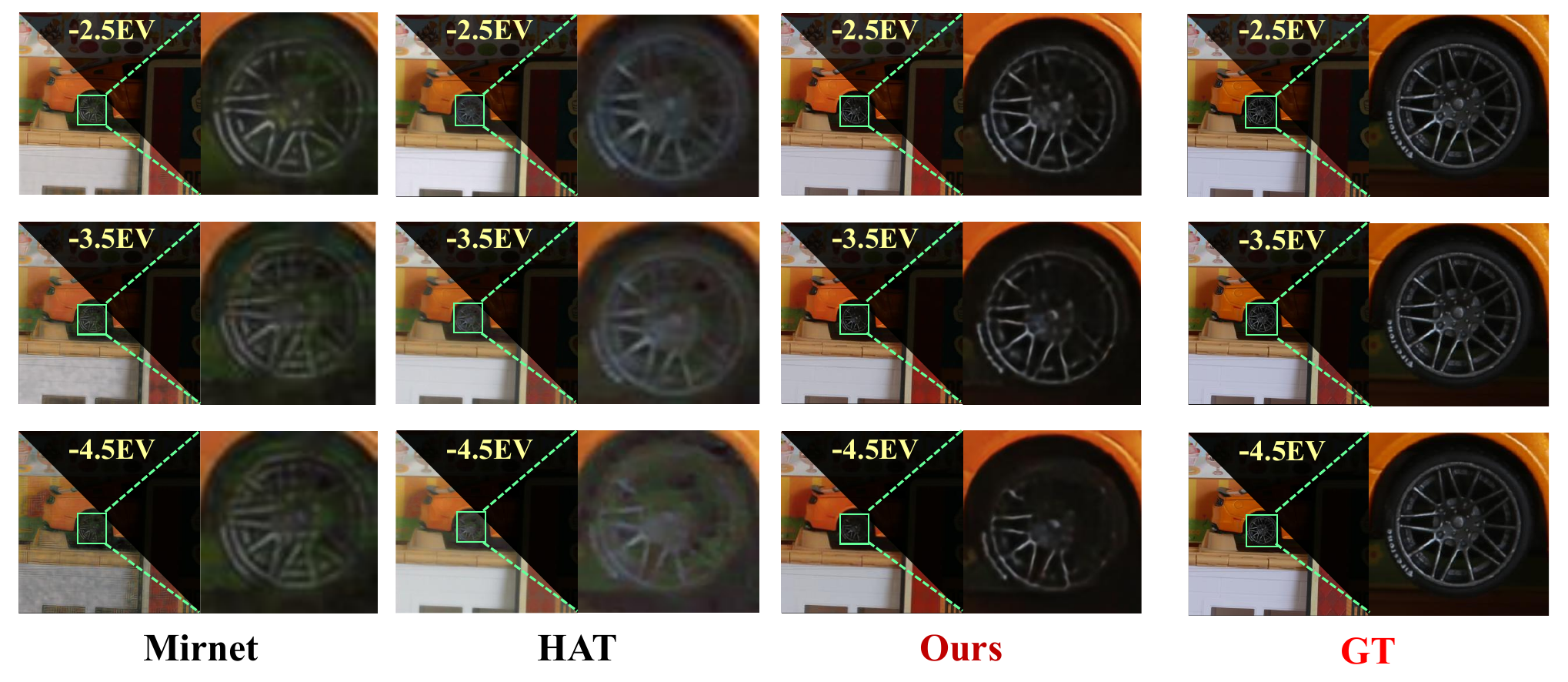}}
    \caption{Visual comparison under extreme low-light conditions ($-2.5$ EV to $-4.5$ EV) on the RELLISUR dataset.}
    \label{fig_different_light_compare}
\vspace{-10pt}
\end{figure}

\subsection{Cross-frequency Semantic Recomposition }

To integrate the enhanced luminance and restored texture features into a unified high-fidelity representation\cite{11417419}, we design a \MC module. Instead of naive concatenation, \MC leverages dual-branch attention to perform content-aware, frequency-sensitive feature fusion. By adaptively calibrating the fusion process according to semantic importance, this design ensures perceptual consistency and structural alignment.

Given the outputs from the previous stage, \ie, enhanced luminance $\mathcal{I}^{hlr}$ and denoised texture $\mathcal{T}^{dlr}$, we first apply two parallel attention pathways to assess semantic contributions across spatial and channel dimensions. In particular, spatial attention $\phi_{\text{SA}}(\cdot)$ and channel attention $\phi_{\text{CA}}(\cdot)$ are computed to capture complementary fusion cues. Their contributions are then balanced by a learnable gating function $\mathcal{G}_{\text{att}} \in [0, 1]$, which produces a soft selection mask based on input dynamics and can be represented as:
\begin{equation}
\mathcal{F}^{csr} = \mathcal{G}_{att} \cdot \phi_{CA}(\mathcal{I}^{hlr}, \mathcal{T}^{dlr}) + (1 - \mathcal{G}_{att}) \cdot \phi_{SA}(\mathcal{I}^{hlr}, \mathcal{T}^{dlr}).
\end{equation}

To further mitigate information degradation during fusion, we design a residual refinement path from $\mathcal{T}^{dlr}$, reinforcing high-frequency structure retention. The final high-resolution output $\mathcal{R}^{hsr}$ is reconstructed through a lightweight decoder consisting of feature mixing, structural refinement, and resolution upsampling, which can be denoted as:
\begin{equation}
\mathcal{R}^{hsr} = \text{Upsample}(\text{Rebuild}(\mathcal{F}^{csr} + \mathcal{T}^{dlr})).
\end{equation}

By explicitly aligning cross-frequency representations and dynamically adjusting fusion priorities, \MC effectively reconciles global luminance coherence with local texture fidelity. This ensures that the reconstructed image maintains consistent structure and perceptual realism under diverse low-light degradation patterns.

\begin{table*}[t]
    \centering
    \small
    \caption{Comparison of our \name~with sota methods on RELLISUR dataset at $\times$2 and $\times$4 scales. (\textbf{Bold}: Best, \underline{Underline}: Second best)}
    \renewcommand{\arraystretch}{1.25}
    \setlength{\tabcolsep}{10pt}
    \begin{tabular}{cccccccc}
    \hline
    \multirow{2}{*}{Method} & \multirow{2}{*}{Years} & \multicolumn{3}{c}{RELLISUR $\times$2} & \multicolumn{3}{c}{RELLISUR $\times$4} \\
    \cline{3-8}
    & & PSNR$\uparrow$ & SSIM$\uparrow$ & LPIPS$\downarrow$ & PSNR$\uparrow$ & SSIM$\uparrow$ & LPIPS$\downarrow$ \\
    \hline
    
      
    SwinIR~\cite{liang2021swinir}     & ICCV'21  & 18.38$_{\uparrow4.77}$ & 0.640$_{\uparrow0.176}$ & 0.577$_{\downarrow0.453}$ & 17.53$_{\uparrow4.33}$ & 0.663$_{\uparrow0.151}$ & 0.688$_{\downarrow0.435}$ \\
    Restormer~\cite{zamir2022restormer}  & CVPR'22  & 21.21$_{\uparrow1.94}$ & 0.727$_{\uparrow0.089}$ & 0.385$_{\downarrow0.261}$ & 20.29$_{\uparrow1.57}$ & 0.720$_{\uparrow0.094}$ & 0.492$_{\downarrow0.239}$ \\
    OmniSR~\cite{wang2023omni}  & CVPR'23  & 20.87$_{\uparrow2.28}$ & 0.740$_{\uparrow0.076}$ & 0.470$_{\downarrow0.346}$ & 19.89$_{\uparrow1.97}$ & 0.723$_{\uparrow0.091}$ & 0.461$_{\downarrow0.208}$ \\
    HAT\cite{chen2023activating}        & CVPR'23  & 20.21$_{\uparrow2.94}$ & 0.719$_{\uparrow0.097}$ & 0.454$_{\downarrow0.330}$ & 19.75$_{\uparrow2.11}$ & 0.715$_{\uparrow0.099}$ & 0.561$_{\downarrow0.308}$ \\

    PDHAT+\cite{li2024perceptual}   & T-MM'24 & - &
    - & -  & 21.00$_{\uparrow0.86}$ & \underline{0.753}$_{\uparrow0.061}$ & 0.452$_{\downarrow0.159}$\\
    CollaBA~\cite{gao2024collaborative}   & PR'24 & 21.62$_{\uparrow1.53}$ &
    \underline{0.787}$_{\uparrow0.029}$ & 0.247$_{\downarrow0.123}$& 20.42$_{\uparrow1.44}$ & 0.734$_{\uparrow0.080}$ & \underline{0.371}$_{\downarrow0.118}$\\
    TriCo~\cite{gao2024enhancing}  & ACM MM'24 & 22.45$_{\uparrow0.70}$ &
    0.744$_{\uparrow0.072}$ & 0.304$_{\downarrow0.180}$& 21.05$_{\uparrow0.81}$ & 0.731$_{\uparrow0.083}$ & 0.432$_{\downarrow0.179}$\\
    UltraIS~\cite{gao2024dual}  & T-NNLS'24 & 22.26$_{\uparrow0.89}$ &
    0.743$_{\uparrow0.073}$ & 0.254$_{\downarrow0.130}$& 21.03$_{\uparrow0.83}$ & 0.726$_{\uparrow0.088}$ & 0.371$_{\downarrow0.118}$\\
    MSIRNet~\cite{ye2024learning}    & IF'24 & 21.33$_{\uparrow1.82}$ & 0.750$_{\uparrow0.066}$ & 0.365$_{\downarrow0.241}$ & 20.32$_{\uparrow1.54}$ & 0.735$_{\uparrow0.079}$ & 0.437$_{\downarrow0.184}$ \\
    BrZoNet~\cite{yue2024unveiling}    & AAAI'24  & \underline{22.79}$_{\uparrow0.36}$ & 0.745$_{\uparrow0.071}$ & \underline{0.243}$_{\downarrow0.119}$ & \underline{21.41}$_{\uparrow0.45}$ & 0.726$_{\uparrow0.088}$ & 0.383$_{\downarrow0.130}$ \\
    \hline
    \textbf{\name (Ours)} & -    & \textbf{23.15} & \textbf{0.816} & \textbf{0.124} & \textbf{21.86} & \textbf{0.814} & \textbf{0.253} \\
    \hline
    \end{tabular}
    \label{tab:quantitative}
\\[-2ex]
\end{table*}

\begin{table}[t]
    \centering
    \caption{Ablation study of \MA, \MB, and \MC~modules in \name across $\times2$ and $\times4$ upscaling factors.}
    \renewcommand{\arraystretch}{1.55}
    \setlength{\tabcolsep}{3pt}  
    \scalebox{0.95}{
    \begin{tabular}{>{\centering\arraybackslash}p{0.7cm}
                    >{\centering\arraybackslash}p{0.7cm}
                    >{\centering\arraybackslash}p{0.7cm}
                    >{\centering\arraybackslash}p{0.7cm}
                    >{\centering\arraybackslash}p{1.5cm}
                    >{\centering\arraybackslash}p{1.5cm}
                    >{\centering\arraybackslash}p{1.6cm}}
    \hline
    \multirow{2}{*}{Scale} & \multicolumn{3}{c}{Modules} & \multicolumn{3}{c}{Metric} \\
    \cline{2-7}
     & FSD & SDR & CSR & PSNR$\uparrow$ & SSIM$\uparrow$ & LPIPS$\downarrow$ \\
    \hline
    \multirow{8}{*}{$\times2$}
     & $\times$ & $\times$ & $\times$ & 18.51 & 0.765 & 0.387 \\\cline{2-7}
     & \checkmark & $\times$ & $\times$ & 22.36$_{\uparrow3.85}$ & 0.803$_{\uparrow0.038}$ & 0.291$_{\downarrow0.096}$ \\
     & $\times$ & \checkmark & $\times$ & 20.27$_{\uparrow1.76}$ & 0.782$_{\uparrow0.017}$ & 0.362$_{\downarrow0.025}$ \\
     & $\times$ & $\times$ & \checkmark & 21.79$_{\uparrow3.28}$ & 0.791$_{\uparrow0.026}$ & 0.313$_{\downarrow0.074}$ \\\cline{2-7}
     & \checkmark & \checkmark & $\times$ & 22.75$_{\uparrow4.24}$ & 0.811$_{\uparrow0.046}$ & 0.203$_{\downarrow0.184}$ \\
     & \checkmark & $\times$ & \checkmark & 22.73$_{\uparrow4.22}$ & 0.813$_{\uparrow0.046}$ & 0.185$_{\downarrow0.202}$ \\
     & $\times$ & \checkmark & \checkmark & 22.13$_{\uparrow3.62}$ & 0.809$_{\uparrow0.046}$ & 0.239$_{\downarrow0.148}$ \\\cline{2-7}
     & \checkmark & \checkmark & \checkmark & \textbf{23.15}$_{\uparrow4.64}$ & \textbf{0.817}$_{\uparrow0.052}$ & \textbf{0.124}$_{\downarrow0.263}$ \\
    \hline
    \multirow{8}{*}{$\times4$}
     & $\times$ & $\times$ & $\times$ & 17.72 & 0.733 & 0.479 \\\cline{2-7}
     & \checkmark & $\times$ & $\times$ & 21.11$_{\uparrow3.39}$ & 0.801$_{\uparrow0.068}$ & 0.393$_{\downarrow0.086}$ \\
     & $\times$ & \checkmark & $\times$ & 19.41$_{\uparrow1.69}$ & 0.763$_{\uparrow0.030}$ & 0.420$_{\downarrow0.059}$ \\
     & $\times$ & $\times$ & \checkmark & 20.03$_{\uparrow2.31}$ & 0.787$_{\uparrow0.054}$ & 0.407$_{\downarrow0.072}$ \\\cline{2-7}
     & \checkmark & \checkmark & $\times$ & 21.47$_{\uparrow3.75}$ & 0.809$_{\uparrow0.076}$ & 0.326$_{\downarrow0.153}$ \\
     & \checkmark & $\times$ & \checkmark & 21.33$_{\uparrow3.61}$ & 0.811$_{\uparrow0.078}$ & 0.309$_{\downarrow0.170}$ \\
     & $\times$ & \checkmark & \checkmark & 20.79$_{\uparrow3.07}$ & 0.806$_{\uparrow0.073}$ & 0.377$_{\downarrow0.102}$ \\\cline{2-7}
     & \checkmark & \checkmark & \checkmark & \textbf{21.86}$_{\uparrow4.14}$ & \textbf{0.814}$_{\uparrow0.081}$ & \textbf{0.253}$_{\downarrow0.226}$ \\
    \hline
    \end{tabular}}
    \label{tab:ablation}
    \\[-4ex]
\end{table}

\section{Experiments}
\subsection{Experiment Settings}
\subsubsection{Datasets}
We evaluate our proposed method on the RELLISUR dataset~\cite{aakerberg2021rellisur}, a recently curated benchmark for low-light image super-resolution. This dataset provides paired low-resolution dark-light and high-resolution normal-light images at three scaling levels: $\times 1$, $\times 2$, and $\times 4$. Specifically, it contains 3,610 training pairs and 425 test pairs, with the test set encompassing diverse illumination conditions and varying degrees of darkness. To evaluate the robustness of our method under different upscaling demands, we conduct experiments on both the $\times 2$ and $\times 4$ subsets. 


\subsubsection{Validation Metrics}
To comprehensively evaluate the quality of image restoration, we adopt three widely used metrics, Peak Signal-to-Noise Ratio (PSNR), Structural Similarity Index (SSIM), and Learned Perceptual Image Patch Similarity (LPIPS), each capturing different aspects of fidelity and perceptual quality.
These metrics jointly evaluate pixel accuracy, perceptual structure fidelity, and semantic perceptual similarity, offering a comprehensive assessment of restoration quality under challenging low-light conditions.

\subsubsection{Implementation Details}
The specific implementation details and training configurations are provided in the Supplementary Material.

\subsection{Comparison with State-of-the-Artst Methods}
We compare our proposed \name framework against a broad range of state-of-the-art methods, not only including general-purpose image restoration models (\eg, PAN, MIRNet, Restormer), super-resolution models (\eg, SwinIR, HAT, OmniSR), but also containing recent \ML~methods tailored to the dual-degradation setting (\eg, PDHAT+, CollaBA, TriCo). All baselines are retrained under unified protocols on the RELLISUR dataset to ensure a fair comparison.

General image restoration methods, while effective in isolated tasks, often struggle with the compound degradation in \ML~due to the lack of semantic decoupling. Their tendency to process brightness and texture jointly in the spatial domain leads to over-smoothed structures and amplified artifacts under extreme lighting conditions. In contrast, our \name framework explicitly separates these factors in the frequency domain, providing more stable and structurally coherent representations.
Super-resolution approaches are powerful in capturing long-range dependencies and restoring high-frequency details. However, their performance deteriorates under illumination degradation, largely due to their reliance on spatial coupling and the absence of dedicated luminance modeling. Our approach complements frequency-aware decomposition with perceptually driven brightness enhancement, enabling more robust and balanced recovery.
Although SOTA \ML-specific methods attempt to jointly address illumination and resolution to achieve high performance, they rely on complex fusion strategies and lack structured disentanglement of heterogeneous components that inevitably increase computational cost without solving the underlying semantic interference. Instead of using complex structures, our method adopts a streamlined pipeline, \ie, decoupling, specialization, and recomposition that promotes semantic clarity, localized adaptation, and perceptual alignment.


Overall, the results confirm that \name’s structured modeling offers consistent gains across both fidelity and perceptual quality, especially in challenging low-light environments where conventional designs falter.

\subsection{Visualization}
\subsubsection{Qualitative Visualization}
To further validate the perceptual effectiveness of our method under severe low-light degradation, we present qualitative comparisons on the RELLISUR dataset in Fig.~\ref{fig:RGB}. Each sample displays the restored full image (top), a zoomed-in region for local structure inspection (middle), and RGB histogram distributions (bottom). Compared to existing methods, our approach demonstrates superior restoration fidelity both visually and statistically.
Specifically, general SR and \ML~methods often struggle to jointly recover global luminance and fine textures, resulting in either over-smoothed appearance or noisy over-enhancement. In contrast, our model produces visually coherent outputs with clearly restored structural boundaries and sharp local textures. The highlighted patch shows that bottle contours and label text are significantly more legible in our result, indicating better preservation of high-frequency details.
Moreover, the RGB histogram of our reconstruction closely aligns with that of the ground truth, capturing both intensity spread and color consistency. This suggests that our frequency-aware decoupling and cross-semantic fusion contribute not only to spatial structure recovery but also to perceptual alignment across illumination and chrominance domains. The visualization clearly evidences our method’s advantage in addressing dual degradation by the proposed semantically disentangled method and the structurally consistent manner.

\subsubsection{Robustness under Extreme Low-Light Conditions}
To evaluate the robustness of our method across varying levels of illumination degradation, we conduct a stress test on the RELLISUR dataset under extreme low-light settings ranging from $-2.5$ EV to $-4.5$ EV, as shown in Fig.~\ref{fig_different_light_compare}. As the exposure drops, baseline methods progressively lose the ability to recover wheel contours and boundary edges, exhibiting either oversaturation or structure collapse. In contrast, our method retains the geometric layout and spoke sharpness even under $-4.5$ EV, demonstrating superior resilience to noise, blur, and luminance suppression. The zoom-in patches clearly show that our model avoids both over-smoothing and over-enhancement, instead producing coherent reconstructions with natural brightness and preserved detail.
This robustness stems from our frequency-aware decoupling design and semantics-specific enhancement strategy, which isolate and reconstruct luminance and texture components under tailored constraints. The results highlight our model’s ability to adapt to severe degradations while ensuring fine-grained detail recovery and perceptual realism, outperforming conventional approaches across the entire exposure range.

\subsection{Ablation Study}
\subsubsection{Effectiveness of Each Module}
To evaluate the impact of each proposed component and their combined effects, we perform ablation studies by testing each module individually and in various combinations. As shown in Tab.~\ref{tab:ablation}, all modules contribute positively to model performance, with the full \name framework achieving the best results across scales and metrics.
We first analyze the idividual contribution of each module. When a single module is activated, the \MA, \MB, \MC~modules all yield a significant performance boost over the baseline. Specifically, \MA~alone improves the PSNR from $18.51$ dB to $22.36$ dB, hightlighting its effectiveness in decoupling entangled semantics at the input stage. Next, we explore the synergistic effects of combining two modules. The combination of \MA and \MB further enhances the PSNR to $22.75$ dB, which is a notable improvement over either module individually, validating their complementary roles in semantic confusion and disentangled representation. Similarly, pairing \MA with \MC achieves a PSNR of $22.75$ dB, while \MB with \MC results in a PSNR of $22.13$ dB. These results indicate that our modules are designed to work together, with each component reinforcing the others' strengths.
The performance consistently improves as more modules are integrated, which effectively validates our design principle, progressive decoupling, specialization, and recomposition, further enabling more robust and semantically aligned restoration under low-light degradation.

\subsubsection{Effectiveness of Sub-Components in \MB Module}
To explore the core of the \MB module, we further validate the effectiveness of the BioLuma and HicDeno components within the SDR module. The specific quantitative results and analysis are presented in the Supplementary Material.

\section{Conclusion}
To tackle the dual challenges of luminance degradation and texture loss in low-light image super-resolution, we design a frequency-aware restoration framework \name. Unlike conventional approaches that entangle semantic components in spatial-domain modeling, \name explicitly separates luminance and texture into semantically independent frequency subspaces through a frequency-aware structural decoupling process. Based on this disentangled representation, \name employs a semantics-specific dual-path mechanism to perform dedicated luminance enhancement and texture reconstruction, followed by a cross-frequency semantic recomposition module that adaptively fuses heterogeneous cues with spatial and channel attention to ensure perceptual alignment and structural fidelity, respectively. 
Totally, these modules form a unified framework that systematically mitigates semantic confusion and enhances generalization under diverse and extremely low-light conditions. Extensive experiments on public benchmarks validate the superiority of \name on accuracy and recognition rate in terms of both distortion-based and perceptual quality metrics. 
\section{Acknowledgment}
This work is supported by National Natural Science Foundation of China (Grant No. 62302145) and Fundamental Research Funds for the Central Universities (Grant No. JZ2025HGTB0225). 

\bibliographystyle{IEEEbib}
\bibliography{icme2026references}

@inproceedings{liu2022noise,
  title={A noise-aware framework for blind image super-resolution},
  author={Liu, Guanqun and Wang, Xin and Wang, Lei and Zha, Daren and Zhao, Lin and Kong, Zhe and Qi, Peng},
  booktitle={2022 IEEE International Conference on Multimedia and Expo (ICME)},
  pages={01--06},
  year={2022},
  organization={IEEE}
}

@inproceedings{liang2021swinir,
  title={Swinir: Image restoration using swin transformer},
  author={Liang, Jingyun and Cao, Jiezhang and Sun, Guolei and Zhang, Kai and Van Gool, Luc and Timofte, Radu},
  booktitle={Proceedings of the IEEE/CVF international conference on computer vision},
  pages={1833--1844},
  year={2021}
}

@article{li2021learning,
  title={Learning to enhance low-light image via zero-reference deep curve estimation},
  author={Li, Chongyi and Guo, Chunle and Loy, Chen Change},
  journal={IEEE transactions on pattern analysis and machine intelligence},
  volume={44},
  number={8},
  pages={4225--4238},
  year={2021},
  publisher={IEEE}
}

@inproceedings{aakerberg2021rellisur,
  title={RELLISUR: A real low-light image super-resolution dataset},
  author={Aakerberg, Andreas and Nasrollahi, Kamal and Moeslund, Thomas B},
  booktitle={Thirty-fifth Conference on Neural Information Processing Systems-NeurIPS 2021},
  year={2021}
}

@inproceedings{yue2024unveiling,
  title={Unveiling details in the dark: Simultaneous brightening and zooming for low-light image enhancement},
  author={Yue, Ziyu and Gao, Jiaxin and Su, Zhixun},
  booktitle={Proceedings of the AAAI Conference on Artificial Intelligence},
  volume={38},
  number={7},
  pages={6899--6907},
  year={2024}
}

@article{ye2024learning,
  title={Learning multi-granularity semantic interactive representation for joint low-light image enhancement and super-resolution},
  author={Ye, Jing and Liu, Shenghao and Qiu, Changzhen and Zhang, Zhiyong},
  journal={Information Fusion},
  volume={110},
  pages={102467},
  year={2024},
  publisher={Elsevier}
}

@inproceedings{gao2024enhancing,
  title={Enhancing Images with Coupled Low-Resolution and Ultra-Dark Degradations: A Tri-level Learning Framework},
  author={Gao, Jiaxin and Liu, Yaohua},
  booktitle={Proceedings of the 32nd ACM International Conference on Multimedia},
  pages={8642--8651},
  year={2024}
}

@article{li2024perceptual,
  title={Perceptual decoupling with heterogeneous auxiliary tasks for joint low-light image enhancement and deblurring},
  author={Li, Yuezhou and Xu, Rui and Niu, Yuzhen and Guo, Wenzhong and Zhao, Tiesong},
  journal={IEEE Transactions on Multimedia},
  volume={26},
  pages={6663--6675},
  year={2024},
  publisher={IEEE}
}

@article{gao2024collaborative,
  title={Collaborative brightening and amplification of low-light imagery via bi-level adversarial learning},
  author={Gao, Jiaxin and Liu, Yaohua and Yue, Ziyu and Fan, Xin and Liu, Risheng},
  journal={Pattern Recognition},
  volume={154},
  pages={110558},
  year={2024},
  publisher={Elsevier}
}

@inproceedings{chen2023activating,
  title={Activating more pixels in image super-resolution transformer},
  author={Chen, Xiangyu and Wang, Xintao and Zhou, Jiantao and Qiao, Yu and Dong, Chao},
  booktitle={Proceedings of the IEEE/CVF conference on computer vision and pattern recognition},
  pages={22367--22377},
  year={2023}
}

@inproceedings{zamir2022restormer,
  title={Restormer: Efficient transformer for high-resolution image restoration},
  author={Zamir, Syed Waqas and Arora, Aditya and Khan, Salman and Hayat, Munawar and Khan, Fahad Shahbaz and Yang, Ming-Hsuan},
  booktitle={Proceedings of the IEEE/CVF conference on computer vision and pattern recognition},
  pages={5728--5739},
  year={2022}
}

@inproceedings{wang2023omni,
  title={Omni aggregation networks for lightweight image super-resolution},
  author={Wang, Hang and Chen, Xuanhong and Ni, Bingbing and Liu, Yutian and Liu, Jinfan},
  booktitle={Proceedings of the IEEE/CVF Conference on Computer Vision and Pattern Recognition},
  pages={22378--22387},
  year={2023}
}

@article{gao2024dual,
  title={A dual-stream-modulated learning framework for illuminating and super-resolving ultra-dark images},
  author={Gao, Jiaxin and Yue, Ziyu and Liu, Yaohua and Xie, Sihan and Fan, Xin and Liu, Risheng},
  journal={IEEE transactions on neural networks and learning systems},
  year={2024},
  publisher={IEEE}
}

@ARTICLE{11417419,
  author={Huang, Jinyang and Feng, Yuanhao and Cui, Feng-Qi and Zhang, Xiang and Liu, Zhi and Liu, Xin and Liu, Jianchun and Zhang, Fusang and Li, Meng},
  journal={IEEE Transactions on Dependable and Secure Computing}, 
  title={Identifying Who You Are No Matter What You Write through Abstracting Handwriting Style}, 
  year={2026},
  volume={},
  number={},
  pages={1-15},
  doi={10.1109/TDSC.2026.3668275}}

@inproceedings{cui2025learning,
  title={Learning from heterogeneity: Generalizing dynamic facial expression recognition via distributionally robust optimization},
  author={Cui, Feng-Qi and Tong, Anyang and Huang, Jinyang and Zhang, Jie and Guo, Dan and Liu, Zhi and Wang, Meng},
  booktitle={Proceedings of the 33rd ACM international conference on multimedia},
  pages={5587--5596},
  year={2025}
}

\end{document}